\def\paperlanguage{}
\pdfoutput=1 

\documentclass[letterpaper, 10 pt, conference]{ieeeconf}  

\usepackage{bm}
\usepackage{cite}
\usepackage{flushend}
\include{preamble}

\newcommand{\ctext}[1]{\raise0.2ex\hbox{\textcircled{\scriptsize{#1}}}}

\IEEEoverridecommandlockouts                              

\title{\LARGE \textbf
  {
    \switchlanguage%
    {%
      Self-Supervised Learning of Visual Servoing for Low-Rigidity Robots Considering Temporal Body Changes
    }%
    {%
      低剛性ロボットの身体変化を考慮した自律的視覚サーボ学習
    }%
  }
}

\author{Kento Kawaharazuka$^{1}$, Naoaki Kanazawa$^{1}$, Kei Okada$^{1}$, and Masayuki Inaba$^{1}$
  \thanks{$^{1}$ The authors are with the Department of Mechano-Informatics, Graduate School of Information Science and Technology, The University of Tokyo, 7-3-1 Hongo, Bunkyo-ku, Tokyo, 113-8656, Japan.
    {\texttt\small [kawaharazuka, kanazawa, k-okada, inaba]@jsk.t.u-tokyo.ac.jp}
  }
}
\begin{document}

\maketitle
\thispagestyle{empty}
\pagestyle{empty}

\begin{abstract}
  \switchlanguage%
  {%
    In this study, we investigate object grasping by visual servoing in a low-rigidity robot.
    It is difficult for a low-rigidity robot to handle its own body as intended compared to a rigid robot, and calibration between vision and body takes some time.
    In addition, the robot must constantly adapt to changes in its body, such as the change in camera position and change in joints due to aging.
    Therefore, we develop a method for a low-rigidity robot to autonomously learn visual servoing of its body.
    We also develop a mechanism that can adaptively change its visual servoing according to temporal body changes.
    We apply our method to a low-rigidity 6-axis arm, MyCobot, and confirm its effectiveness by conducting object grasping experiments based on visual servoing.
  }%
  {%
    本研究は低剛性なロボットにおける視覚サーボによる物体把持を行う.
    一方問題として, 低剛性なロボットは剛なロボットに対して自身の身体を意図したように扱うことが難しく, 視覚-身体間のキャリブレーションにも時間を要する.
    また, カメラの位置がずれる, 経年劣化により身体が変化する等, その身体変化に常に適応しなければならない.
    そこで本研究では, 低剛性なロボットが身体の視覚サーボ方法を自律的に学習する手法を開発する.
    また, 身体変化に応じてその視覚サーボを適応的に変化可能な機構も開発する.
    本研究を低剛性な6軸アームであるMyCobotに適用し, 視覚サーボに基づく物体把持実験によりその有効性を確認する.
  }%
\end{abstract}

\section{INTRODUCTION}\label{sec:introduction}
\switchlanguage%
{%
  Visual servoing is one of the most basic control strategies in robotics.
  It is usually applied to industrial robots that are highly rigid and have accurately calibrated vision and bodies \cite{shirai1973vfeedback, espiau1992vservoing}.
  These methods often require carefully designed vision features and manually calibrated models.
  On the other hand, in addition to these general visual servoing methods, learning-based methods have been developed recently due to the advancement of deep learning \cite{lampe2013vservoing, lee2017vservoing, zhang2018imitation}.
  There are two main types of learning methods: learning from demonstration \cite{zhang2018imitation, saito2021tooluse}, in which a human teaches the movement, and reinforcement learning \cite{lampe2013vservoing, lee2017vservoing}, in which the control policy is learned from rewards through many trials in simulation.
  While the former is easy to apply to actual robots, it requires a lot of effort for human demonstration.
  The latter is based on learning in simulation, and is challenging to apply to robots that are difficult to modelize in simulation.

  There are only few examples of applying visual servoing to low-rigidity robots whose bodies do not move as intended.
  In order to perform visual servoing on these low-rigidity robots, it is necessary to accurately modelize their flexible bodies and spend time on the calibration between vision and body \cite{wang2017vservoing, chatelain2015vservoing}.
  In addition, the body of a low-rigidity robot tends to change over time, which causes the model to change gradually, requiring multiple iterations of modeling and calibration \cite{kawaharazuka2020dynamics}.
  Note that there are some examples of using a soft gripper for the hand of a rigid robot \cite{choi2018learning}, but this is different from the focus of this study.
}%
{%
  視覚サーボはロボットにおける最も基本的な制御戦略である.
  これは高剛性で視覚・身体が正確にキャリブレーションされた産業用ロボット等に適用されるのが基本である\cite{shirai1973vfeedback, espiau1992vservoing}.
  これらは人間により丁寧に設計された特徴量とキャリブレーションされたカメラを要する場合が多い.
  一方で, これら一般的な視覚サーボだけでなく, 最近は深層学習の発展により, 学習型の視覚サーボ手法が開発されている\cite{lampe2013vservoing, lee2017vservoing, zhang2018imitation}.
  学習型は主に, 人間がその動きを教えるlearning from demonstration型\cite{zhang2018imitation, saito2021tooluse}と, シミュレーションにおける多数の試行によりpolicyを学習する強化学習型\cite{lampe2013vservoing, lee2017vservoing}が存在する.
  前者は実機においても適用が容易である一方で, 人間のdemonstrationが必要な点で労力が必要である.
  後者は, シミュレーションにおける学習が前提であり, シミュレーションの難しいロボットに適用するのは困難である.

  一方で, 身体が意図した通りに動かないような低剛性なロボットへの適用例は多くない.
  これら低剛性なロボットにおいて視覚サーボを行う際には, 正確にその柔軟身体のモデルを立て, 視覚-身体間のキャリブレーションに時間を費やす必要がある\cite{wang2017vservoing, chatelain2015vservoing}.
  また, 低剛性なロボットは経年劣化等によって身体が変化しやすく, これによりモデルは逐次的に変化するため, 何度もモデル化やキャリブレーションを行う必要がある\cite{kawaharazuka2020dynamics}.
  一方で, ロボットのハンド部分のみソフトロボットハンドを用いた例は存在するが, 本研究の趣旨とは異なる\cite{choi2018learning}.
}%

\begin{figure}[t]
  \centering
  \includegraphics[width=0.9\columnwidth]{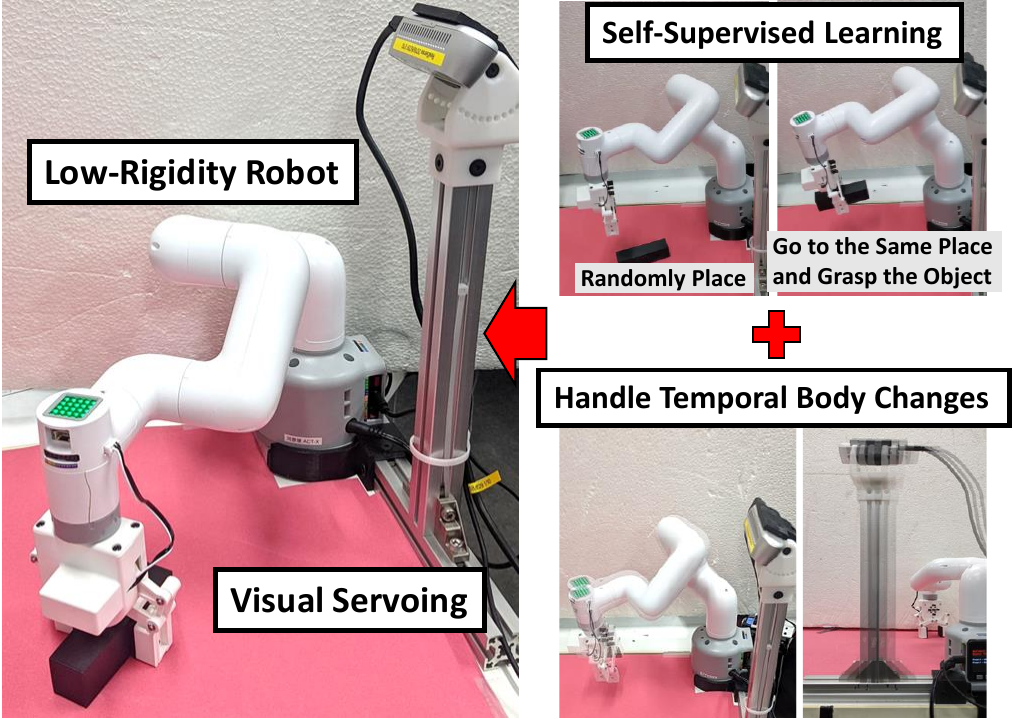}
  \caption{The concept of this study: self-supervised learning of visual servoing and handling of temporal body changes for low-rigidity robots.}
  \label{figure:concept}
\end{figure}

\begin{figure*}[t]
  \centering
  \includegraphics[width=1.99\columnwidth]{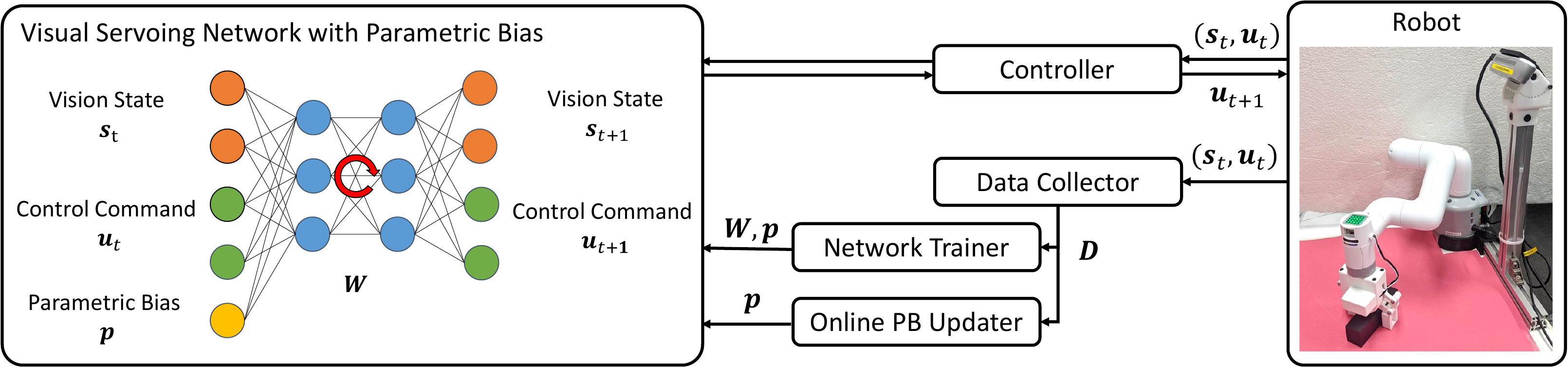}
  \caption{The overall system of visual servoing network with parametric bias.}
  \label{figure:network-structure}
\end{figure*}

\switchlanguage%
{%
  Based on these problems, we propose a method for a robot with a low rigidity and difficult-to-modelize body to learn its own law of visual servoing autonomously, taking into account temporal body changes (\figref{figure:concept}).
  The autonomous data collection is achieved not by human demonstration or reinforcement learning, but by repeatedly placing and picking up an object by itself.
  This is based on the fact that while it is difficult for the robot to directly grasp an object by visual recognition, it can grasp an object by reaching out to the exact same place if the object had been placed by itself.
  There is a similar data collection method \cite{kanamura2021educational}, but it focuses only on automatic data collection for a rigid robot, and its goal is different from this study, which utilizes the motion reproducibility of a low-rigidity robot.
  Although our data collection method is limited to pick-and-place tasks, it is a basic motion common to various tasks, and we believe that it would be useful for cost effective robot that cannot move accurately due to low rigidity to learn such tasks autonomously.
  For adaptation to temporal body changes, we introduce parametric bias \cite{tani2002parametric} and the robot adapts to the current body state by online learning of this variable.
  Parametric bias is a learnable input variable that has been mainly used for imitation learning in the field of cognitive robotics \cite{ogata2005extracting, yokoya2006rnnpb}, which we apply to visual servoing in this study.
  Changes in the dynamics of the model due to changes in the body over time, which are significant in a low-rigidity body, are implicitly embedded in this parametric bias, and the model adapts to the current state by constantly updating this variable to match the prediction of the vision sensor with the actual measured data.
  This study is useful and versatile in that it does not depend on a specific robot body structure and allows learning of visual servoing autonomously.
  Also, autonomous data collection mainly deals with the low-rigidity body of the robot, while parametric bias deals with the temporal changes of the body.
  Therefore, it is possible to apply these two methods separately, for example, to low-rigidity robots without temporal body changes, high-rigidity robots with temporal body changes, etc.

  The contributions of this study are as follows.
  \begin{itemize}
    \item Data collection for autonomous learning of a visual servoing model for a low-rigidity robot.
    \item Recognition and adaptation to changes in the visual servoing model due to temporal body changes.
    \item Acquisition of an object grasping controller by the actual robot with a low rigidity
  \end{itemize}
  The structure of this study is as follows.
  In \secref{sec:proposed}, we describe the network structure, data collection, network training, adaptation to the current body state, and visual servoing control, in order.
  In \secref{sec:experiment}, we describe data collection for a low-rigidity robot, online learning of parametric bias, and visual servoing experiments.
  We discuss the experimental results in \secref{sec:discussion} and conclude in \secref{sec:conclusion}.
}%
{%
  これらの問題点を踏まえ, 本研究では低剛性でモデル化困難な身体を持つロボットが, 身体の時間的変容を考慮可能かつ自律的に自身の視覚サーボ則を学習していく手法を提案する(\figref{figure:concept}).
  自律的な学習については, 人間がデモンストレーションを与えるのでも強化学習によって学習するのでもなく, 自身で物体を置き, これを取ることを繰り返すことで達成する.
  これは, 直接視覚で認識して物体を取ることは難しい一方で, 自分で置いた物体であれば全く同じ場所に手を伸ばすことで物体把持を行うことができるという特徴を利用する.
  同様の形でデータ収集を行う手法\cite{kanamura2021educational}が存在するが, 高剛性なロボットにおける自動的なデータ収集にのみ着目しており, 低剛性な身体の動作再現性を利用した本研究とは狙いが異なる.
  Pick and Placeに限定したデータ収集法ではあるものの, 様々な作業に共通する基本動作であり, 低剛性で正確に動けないロボットが自律的ににこれを学習することは有用だと考えている.
  また, 身体の時間的変容への適応については, Parametric Bias \cite{tani2002parametric}の導入とこの変数のオンライン学習による現在の身体状態への適応を行う.
  Parametric Biasは認知ロボティクスの分野において, 主に模倣学習に利用されてきた学習可能な入力変数であり\cite{ogata2005extracting, yokoya2006rnnpb}, これを視覚サーボに応用する.
  低剛性なロボットにおいて顕著な経年劣化等による身体変化によるモデルのダイナミクス変化をこのParametric Biasに暗黙的に埋め込み, 感覚の予測と実測値を合致させるようにこの変数を常に更新することで現在状態に適応する.
  本研究は特定のロボット身体構造に依存せず, かつ自律的に学習ができる点で有用かつ高い汎用性を持つ.
  また, 自律的な学習は主にロボットの低剛性身体に対処し, Parametric Biasは身体の時間的変容に対処する.
  そのため, 身体変化のない低剛性身体や, 身体変化のある高剛性ロボットなど, これら2つの手法を別々に適用することも可能である.

  本研究のコントリビューションは以下である.
  \begin{itemize}
    \item 低剛性なロボットにおける視覚サーボモデルの自律学習に向けたデータ収集
    \item 身体への時間的変容による視覚サーボモデル変容の認識と適応
    \item 低剛性なロボット実機による物体把持コントローラの獲得
  \end{itemize}
  本研究の構成は以下のようになっている.
  \secref{sec:proposed}ではネットワーク構成, データ収集, ネットワーク訓練, 現在の身体状態への適応, 視覚サーボ制御について順に述べる.
  \secref{sec:experiment}では低剛性なロボット実機におけるデータ収集, Parametric Biasのオンライン学習, 視覚サーボ実験について述べる.
  \secref{sec:discussion}で実験について考察を述べ, \secref{sec:conclusion}で結論を述べる.
}%

\section{Self-Supervised Learning of Visual Servoing Considering Temporal Body Changes} \label{sec:proposed}
\switchlanguage%
{%
  The overall system of this study is shown in \figref{figure:network-structure}.
  The network proposed in this study is called Visual Servoing Network with Parametric Bias (VSNPB).
}%
{%
  本研究の全体システムの概要を\figref{figure:network-structure}に示す.
  本研究で提案するネットワークをVisual Servoing Network with Parametric Bias (VSNPB)と呼ぶ.
}%

\subsection{Network Structure of VSNPB} \label{subsec:network-structure}
\switchlanguage%
{%
  The network structure of VSNPB can be expressed by the following equation,
  \begin{align}
    (\bm{s}_{t+1}, \bm{u}_{t+1}) = \bm{h}(\bm{s}_{t}, \bm{u}_{t}, \bm{p}) \label{eq:vfnpb}
  \end{align}
  where $t$ is the current time step, $\bm{s}$ is the visual information of the robot, $\bm{u}$ is the robot control input, $\bm{p}$ is the parametric bias (PB), and $\bm{h}$ is the visual servoing model.
  In this study, $\bm{s}$ is the vector $\bm{z}$ where image $I_{t}$ is compressed by AutoEncoder \cite{hinton2006reducing}.
  Also, $\bm{u}$ is the target joint angle of the robot $\bm{\theta}^{ref}$.
  This varies depending on the robot, and for other robots such as tendon-driven robots, the target muscle length $\bm{l}^{ref}$ can be used for $\bm{s}$ \cite{kawaharazuka2019musashi}.
  Parametric bias $\bm{p}$ is a learnable input variable, and by collecting data while changing the body state in various ways, it is possible to embed the change in dynamics of $\bm{h}$ due to temporal body changes into $\bm{p}$.
  Note that parametric bias can only handle static body changes, not dynamic ones.

  In this study, VSNPB has ten layers, consisting of four FC layers (fully-connected layers), two LSTM layers (long short-term memory layers \cite{hochreiter1997lstm}), and four FC layers, in order.
  The number of units is set to \{$N_s+N_u+N_p$, 300, 150, 50, 50 (number of units in LSTM), 50 (number of units in LSTM), 50, 150, 300, $N_s+N_u$\} (where $N_{\{s, u, p\}}$ is the dimensionality of $\{\bm{s}, \bm{u}, \bm{p}\}$).
  The activation function is Hyperbolic Tangent and the update rule is Adam \cite{kingma2015adam}.
  Note that the input and output of the network are normalized using all the collected data.
  Regarding image compression, for a color image of $128\times96$, convolutional layers with kernel size of 3 and stride of 2 are applied five times, the dimensionality is reduced stepwise to 256 and 15 units by two fully-connected layers, and then the image is reconstructed by fully-connected layers and deconvolutional layers in the same way.
  Batch Normalization \cite{ioffe2015batchnorm} is applied to all layers except the last layer, the activation function is ReLU \cite{nair2010relu} for all layers except the last layer (where Sigmoid is used), and the update rule is Adam \cite{kingma2015adam}.
  $\bm{p}$ is two-dimensional and the execution period of \equref{eq:vfnpb} is set to 1 Hz.

  As for the dimension of PB, if it is too large, the space of PB is not self-organized well, and if it is too small, it becomes difficult to consider the temporal body changes.
  In general, it is preferable to set the dimension of PB to be slightly smaller than the dimension of possible changes.
  Even if there are many factors of temporal body changes, setting PB to a smaller dimension allows us to consider body changes better than if PB is not included, and similar body changes are compactly embedded in the same dimension.
}%
{%
  VSNPBのネットワークは以下のように式で表現できる.
  \begin{align}
    (\bm{s}_{t+1}, \bm{u}_{t+1}) = \bm{h}(\bm{s}_{t}, \bm{u}_{t}, \bm{p}) \label{eq:vfnpb}
  \end{align}
  ここで, $t$は現在のタイムステップ, $\bm{s}$はロボットの視覚情報, $\bm{u}$はロボット制御入力, $\bm{p}$はParametric Bias (PB), $\bm{h}$は視覚サーボモデルを表す.
  本研究では$\bm{s}$を, AutoEncoder \cite{hinton2006reducing}によって画像$I_{t}$を圧縮したベクトル$\bm{z}$とする.
  また, $\bm{u}$はロボットの指令関節角度$\bm{\theta}^{ref}$とする.
  これはロボットによっても異なり, 腱駆動ロボットであれば指令筋長$\bm{l}^{ref}$等を用いれば良い\cite{kawaharazuka2019musashi}.
  Parametric Bias $\bm{p}$は学習可能な入力変数であり, 身体状態を様々に変化させながらデータを取ることで, これらによる$\bm{h}$のダイナミクス変化の情報を$\bm{p}$に埋め込むことが可能である.
  なお, Parametric Biasは静的な変化しか扱えない.

  本研究ではVSNPBは10層としており, 順に4つのFC層(fully-connected layer), 2層のLSTM層(long short-term memory layer \cite{hochreiter1997lstm}), 4層のFC層からなる.
  ユニット数については, \{$N_s+N_u+N_p$, 300, 150, 50, 50 (LSTMのunit数), 50 (LSTMのunit数), 50, 150, 300, $N_s+N_u$\}とした(なお, $N_{\{s, u, p\}}$は$\{\bm{s}, \bm{u}, \bm{p}\}$の次元数とする).
  活性化関数はHyperbolic Tangent, 更新則はAdam \cite{kingma2015adam}とする.
  なお, ネットワークの入力と出力は訓練時に得られたデータを使って正規化されている.
  また, 画像を圧縮する際は, $128\times96$のカラー画像について, カーネルサイズが3, ストライドが2の畳み込み層を5回適用し, 全結合層で順にユニット数256, 15まで次元を削減したあと, 同様に全結合層・逆畳み込み層によって画像を復元していく形を取っている.
  最終層以外についてはBatch Normalization \cite{ioffe2015batchnorm}が適用され, 活性化関数は最終層以外についてはReLU \cite{nair2010relu}, 最終層はSigmoid, 更新則はAdam \cite{kingma2015adam}とした.
  また, $\bm{p}$は2次元, \equref{eq:vfnpb}の実行周期は1 Hzとしている.

  PBの次元の決定法であるが, $\bm{p}$は次元を大きくし過ぎると, PBの空間がうまく自己組織化せず, 小さくし過ぎると, 身体変化等を考慮することが難しくなる.
  基本的に, 想定され得る変化の次元よりも少し小さめに設定することが好ましいと考えている.
  例え身体変化の要因が大量にあったとしても, PBを小さめの次元に設定しておくことで, 入れない場合よりは身体変化を考慮できるうえ, 似た身体変化はコンパクトに同じ次元に埋め込まれる.
}%

\subsection{Data Collection} \label{subsec:data-collection}
\switchlanguage%
{%
  In this study, we describe the data collection necessary for learning visual servoing autonomously.
  A low-rigidity robot cannot grasp an object simply by recognizing it and reaching for it as intended due to the nature of the robot.
  On the other hand, if the motion itself is reproducible, the robot can reach to the exact same position as before, when the trajectory of the joint angle is the same.
  The data collection in this study takes advantage of this feature, and the procedure is shown below.
  \renewcommand{\labelenumi}{(\arabic{enumi})}
  \begin{enumerate}
    \item Set the robot to the initial posture.
    \item Have the robot hand grasp the target object in the desired way.
    \item Have the robot reach out to a random position and release the object to return to the initial posture.
    \item Have the robot reach out to the exact same position as before using the same procedure as in (3), grasp the object, and return to the initial posture.
    \item Repeat (3) and (4) until the data collection is complete, then return to (1) and repeat the process of grasping another object.
  \end{enumerate}
  We need to reflect the desired angle and position of the grasp in (2) because this grasping configuration will always be used during the data collection.
  In (3), various motion trajectories are considered, and the robot is moved by the same procedure in (4), so if there is a specific speed of the robot or some specific motions such as pregrasp are required, they should be included here.
  In order to ensure the reproducibility of the robot motion in (4), it is very important to follow the exact same procedure as in (3).
  In particular, if there are dynamic elements in the motion, the robot may reach to different places if the speed of the motion is not kept the same.
  On the other hand, if there is reproducibility, procedures (3) and (4) will always succeed, and the robot can continue to collect data autonomously.
  In this way, it is possible to collect a time series of data as if the robot is executing feedback control of its body to the target object using its vision.
}%
{%
  本研究で自律的に視覚サーボを学習するために必要なデータ収集について述べる.
  低剛性なロボットはその性質から単純に物体を認識してそこに手を伸ばしても思ったとおりに物体を把持できるわけではない.
  一方で, その動き自体に再現性がある場合, 一度手を伸ばしたところに対して, 全く同じ関節角度であれば全く同じ位置に手を伸ばすことができる.
  本研究のデータ収集はこの性質を存分に利用しており, その手順を以下に示す.
  \renewcommand{\labelenumi}{(\arabic{enumi})}
  \begin{enumerate}
    \item ロボットを初期姿勢にする.
    \item 人間がロボットの手に, 所望の物体を把持して欲しい所望の形で把持させる.
    \item ランダムな位置に手を伸ばして物体を離し初期姿勢に戻る.
    \item 先と全く同じ位置に(3)における手続きと全く同じ手順で手を伸ばして物体を把持し初期姿勢に戻る.
    \item (3)と(4)を繰り返しデータ収集が終わったら(1)に戻り別の物体を把持させることを繰り返す.
  \end{enumerate}
  (2)の際における把持形態で常にデータ収集するため, 把持して欲しい角度や位置をここで反映させる必要がある.
  (3)の際は動作軌道に様々なバリエーションが考えられ, (4)ではこれと全く同じ手続きでロボットを動かすため, ここでロボットの動作速度やプレグラスプなどの予備動作等が必要な場合はこれを入れ込む.
  (4)ではロボットの再現性を担保させるため, (3)と全く同じ手続きで動かすという点は重要である.
  特に動的な要素がある場合は速度等が同じにならなければ違う場所に手を伸ばしてしまう可能性がある.
  一方, 再現性が取れればこの(3)と(4)の手順は必ず成功するため, ロボットが自律的にデータを取得し続けることができる.
  これにより, 擬似的にロボットが物体に対して視覚を使い身体をフィードバックするような時系列のデータを収集することが可能である.
}%

\subsection{Training of VSNPB} \label{subsec:network-training}
\switchlanguage%
{%
  Using the obtained data $D$, we train VSNPB.
  In this process, by collecting data while changing the body state, we can implicitly embed this information into the parametric bias.
  When taking into account the changes in the body due to aging, data should be collected for each degree of aging.
  The data collected for the same body state $k$ can be expressed as $D_{k}=\{(\bm{s}_{1}, \bm{u}_{1}), (\bm{s}_{2}, \bm{u}_{2}), \cdots, (\bm{s}_{T_{k}}, \bm{u}_{T_{k}})\}$ ($1 \leq k \leq K$, where $K$ is the total number of body states and $T_{k}$ is the number of motion steps for the trial in the body state $k$), and the data used for training $D_{train}=\{(D_{1}, \bm{p}_{1}), (D_{2}, \bm{p}_{2}), \cdots, (D_{K}, \bm{p}_{K})\}$ is obtained.
  Here, $\bm{p}_{k}$ is the parametric bias that represents the dynamics in the body state $k$, which is a common variable for that state and a different variable for another state.
  Using this $D_{train}$, VSNPB is trained.
  In usual learning processes, only the network weight $W$ is updated, but here, $W$ and $p_{k}$ for each state are updated simultaneously.
  In this way, $p_{k}$ embeds the difference of dynamics in each body state.
  The mean squared error for $\bm{s}$ and $\bm{u}$ is used as the loss function in the training, and all $\bm{p}_{k}$ are optimized with $\bm{0}$ as initial values.
}%
{%
  得られたデータ$D$を使いVSNPBを学習させる.
  この際, 身体状態を変化させながらデータを収集することで, これらのデータを暗黙的にParametric Biasに埋め込むことができる.
  もし経年劣化等による身体変化を考慮したい場合は, 経年劣化の度合いごとにデータを収集していけば良い.
  ある同一の身体状態$k$において収集されたデータを$D_{k}=\{(\bm{s}_{1}, \bm{u}_{1}), (\bm{s}_{2}, \bm{u}_{2}), \cdots, (\bm{s}_{T_{k}}, \bm{u}_{T_{k}})\}$ ($1 \leq k \leq K$, $K$は全試行回数, $T_{k}$はその身体状態$k$における試行の動作ステップ数)として, 学習に用いるデータ$D_{train}=\{(D_{1}, \bm{p}_{1}), (D_{2}, \bm{p}_{2}), \cdots, (D_{K}, \bm{p}_{K})\}$を得る.
  ここで, $\bm{p}_{k}$はその身体状態$k$におけるダイナミクスを表現するParametric Biasであり, その状態については共通の変数, 別の状態については別の変数となる.
  この$D_{train}$を使いVSNPBを学習させる.
  通常の学習ではネットワークの重み$W$のみが更新されるが, ここでは$W$と各状態に関する$p_{k}$が同時に更新される.
  これにより, $p_{k}$にそれぞれの身体状態におけるダイナミクスの違いが埋め込まれることになる.
  学習の際は損失関数として$\bm{s}$と$\bm{u}$に関する平均二乗誤差を使い, $\bm{p}_{k}$は全て$\bm{0}$を初期値として最適化される.
}%

\subsection{Online Update of Parametric Bias} \label{subsec:online-update}
\switchlanguage%
{%
  Using the data $D$ obtained in the current body state, we update the parametric bias online.
  If the network weight $W$ is updated, VSNPB may overfit to the data, but if only the low-dimensional parametric bias $\bm{p}$ is updated, no overfitting occurs.
  This online learning allows us to obtain a model that is always adapted to the current body state.

  Let the number of data obtained be $N^{online}_{data}$, and start online learning when the number of data exceeds $N^{online}_{thre}$.
  For each new data, we set the number of batches as $N^{online}_{batch}$, the number of epochs as $N^{online}_{epoch}$, and the update rule as MomentumSGD.
  For the loss function, we use the mean squared error for $\bm{s}$ only.
  It is also important to note that $W$ is fixed and only $\bm{p}$ is updated.
  Data exceeding $N^{online}_{max}$ are deleted from the oldest ones.
}%
{%
  現在の身体状態において得られたデータ$D$を使い, オンラインでParametric Biasを更新する.
  ネットワークの重み$W$を更新してしまうとVSNPBがそのデータに過学習してしまう可能性があるが, 低次元のParametric Bias $\bm{p}$のみを更新するのであれば過学習は起こらない.
  このオンライン学習により, 常に現在の身体状態に適応したモデルを得ることができる.

  得られたデータ数を$N^{online}_{data}$として, データ数が$N^{online}_{thre}$を超えたところからオンライン学習を始める.
  新しいデータが入るたびにバッチ数を$N^{online}_{batch}$, エポック数を$N^{online}_{epoch}$, 更新則をMomentumSGDとして学習を行う.
  この際, 損失関数は$\bm{s}$のみに関する平均二乗誤差を用いる.
  また, $W$は固定し, $\bm{p}$のみを更新する点が重要である.
  $N^{online}_{max}$を超えたデータは古いものから削除していく.
}%

\subsection{Visual Servoing Using VSNPB} \label{subsec:visual-servoing}
\switchlanguage%
{%
  Visual servoing using VSNPB is very simple.
  We only need to calculate the control input $\bm{u}_{t+1}$ at the next time step by forward propagation from the current visual state $\bm{s}_{t}$ and the control input $\bm{u}_{t}$.
  In this case, $W$ is obtained from \secref{subsec:network-training} and $\bm{p}$ is obtained from \secref{subsec:online-update}.
}%
{%
  VSNPBを使った視覚サーボは非常に単純である.
  現在の視覚状態$\bm{s}_{t}$と制御入力$\bm{u}_{t}$から順伝播により次時刻の制御入力$\bm{u}_{t+1}$を計算するのみである.
  この際, $W$は\secref{subsec:network-training}から得られたもの, $\bm{p}$は\secref{subsec:online-update}から得られたものを使う.
}%

\begin{figure}[t]
  \centering
  \includegraphics[width=0.8\columnwidth]{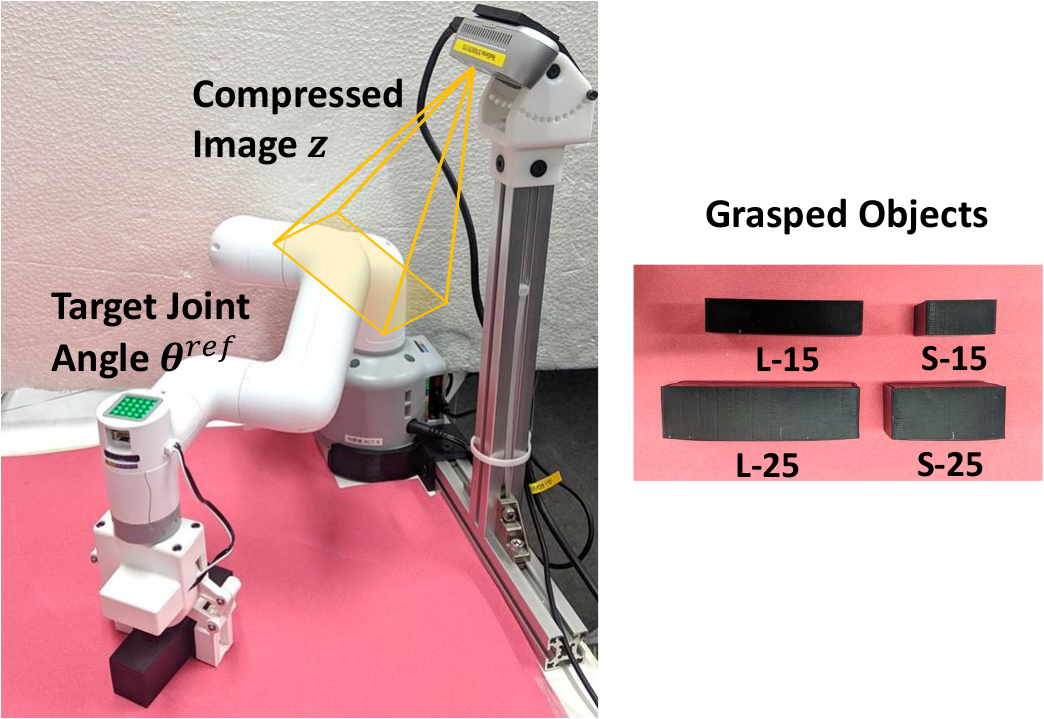}
  \caption{The experimental setup of the low-rigidity robot MyCobot and four grasped objects.}
  \label{figure:exp-setup1}
\end{figure}

\begin{figure}[t]
  \centering
  \includegraphics[width=0.8\columnwidth]{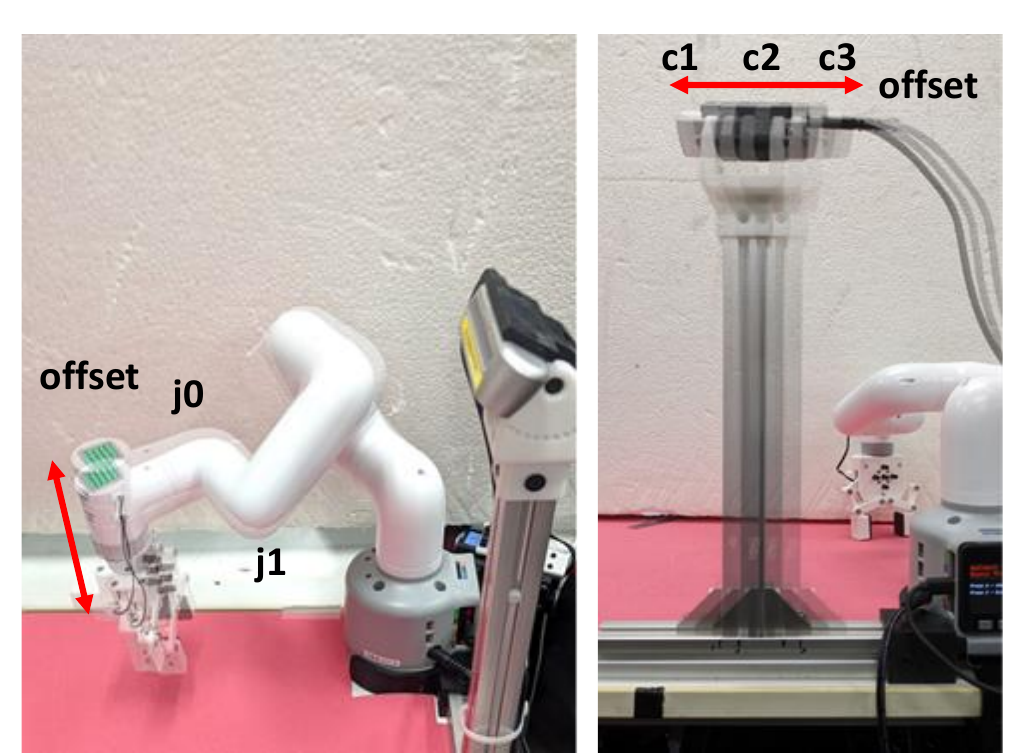}
  \caption{The temporal body changes handled in this study: change in realization of joint angle and change in camera position.}
  \label{figure:exp-setup2}
\end{figure}

\begin{figure*}[t]
  \centering
  \includegraphics[width=2.0\columnwidth]{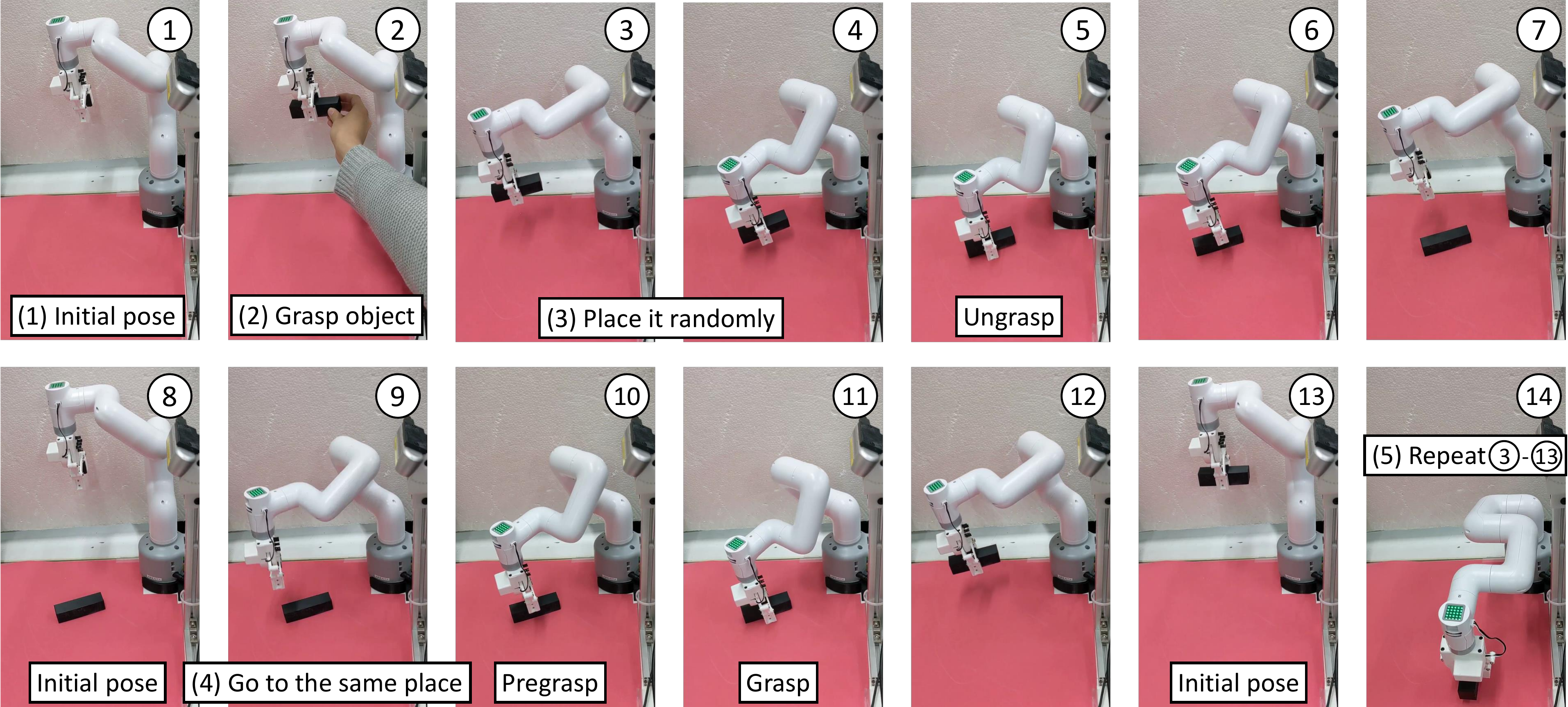}
  \caption{The procedure of data collection for visual servoing.}
  \label{figure:data-collection}
\end{figure*}

\section{Experiments} \label{sec:experiment}

\subsection{Experimental Setup}
\switchlanguage%
{%
  The experimental setup of this study is shown in \figref{figure:exp-setup1}.
  The robot used in this study is MyCobot, whose body is made of plastic and whose links flex and joints easily change due to loosening of screws.
  The camera is RealSense D435.
  $\bm{s}$ is the color image compressed by AutoEncoder $\bm{z}$, and $\bm{u}$ is the target joint angle $\bm{\theta}^{ref}$.
  Here, $\bm{\theta}^{ref}$ is 7-dimensional, which are the dimensions of the six joints of the robot and the opening and closing of the gripper (0 when open and 1 when closed).
  When $\bm{u}$ calculated in \secref{subsec:visual-servoing} is sent to the robot, the gripper is closed if the value is more than 0.5, and open otherwise.
  In this study, we use four objects of different lengths (L or S) and widths (15 mm or 25 mm), denoted L-15, S-15, L-25 and S-25, as shown in \figref{figure:exp-setup1}.
  Note that the maximum opening of the gripper is 38 mm.

  For temporal body changes, we handle the change in the realization of the joint angle and the change in the camera position, as shown in \figref{figure:exp-setup2}.
  This is because the two main aging factors of the low-rigidity robot handled in this study are the misalignment of the origin of joints due to loosening of screws and the misalignment of the camera and the robot due to reassembling of the robot.
  Our experimental setup simply imitates these factors.
  Although it would be desirable to directly handle the temporal changes in the body state while the robot is moving for a long period of time, it is difficult to evaluate the changes quantitatively, so the changes in the body state are virtually created in this study.
  For the realization of the joint angle, we prepare two states, j0 and j1, where j0 is the case that $\bm{\theta}^{ref}$ is directly sent to the robot and j1 is the case that an offset of 2 deg is added to the target angles of all joints.
  For the change of the camera position, c1, c2, and c3 are prepared, and the camera position is offset by 10 mm each.
  Note that it is not necessary to know these offset values in the training of VSNPB, so it is possible to embed body changes that are difficult to explicitly parameterize into PB.
}%
{%
  本研究の実験セットアップを\figref{figure:exp-setup1}に示す.
  使用するロボットは身体がプラスチックで作られたMyCobotであり, 身体が撓むと同時にネジの緩みによって関節の動きが変化しやすい.
  カメラはRealSense D435を用いている.
  $\bm{s}$はカメラから得られたカラー画像をAutoEncoderにより圧縮した$\bm{z}$, $\bm{u}$は関節角度指令値$\bm{\theta}^{ref}$である.
  ここで, $\bm{\theta}^{ref}$は7次元であり, ロボットの6つの関節とグリッパの開閉(開いている時0, 閉じている時1)の次元を持つ.
  \secref{subsec:visual-servoing}において計算された$\bm{u}$をロボットに送る場合, グリッパについては, 値が0.5以上であればグリッパを閉じ, それ以外の場合はグリッパを開けている.
  また, 本研究では\figref{figure:exp-setup1}に示す, 長さ(LまたはS)と幅(15 mmまたは25 mm)の違う4つの物体L-15, S-15, L-25, S-25を用いる.
  なお, gripperの最大開きは38 mmである.

  次に, 身体状態変化としては, \figref{figure:exp-setup2}に示す関節角度の実現度の変化とカメラ位置の変化を扱う.
  これは, これまで実験して感じた, 本研究で扱う低剛性ロボットの主な劣化要因が, ネジの緩みによる関節原点のズレと, 組み立て直しによるカメラとロボットの位置ズレの2つであったからである.
  本研究ではこれらを簡易的に模倣している.
  身体状態変化は長期間ロボットを動かす間の時間変化を直接扱うことが望ましいが, 定量的な評価が困難であるため本研究では擬似的に身体状態変化を作り出している.
  関節角度の実現度については, j0とj1の状態を用意し, j0は$\bm{\theta}^{ref}$を直接ロボットに送る場合, j1は全関節の指令角度に対して2 degのオフセットをのせた場合である.
  また, カメラ位置の変化については, c1, c2, c3を用意し, それぞれはカメラの位置が10 mmずつオフセットしている.
  なお, VSNPBの学習において, データ収集時にこれらのoffset valueを得ておく必要はないため, 本実験のように明示的なパラメータ化が難しい身体変化もPBに埋め込むことが可能である.
}%

\begin{figure}[t]
  \centering
  \includegraphics[width=0.8\columnwidth]{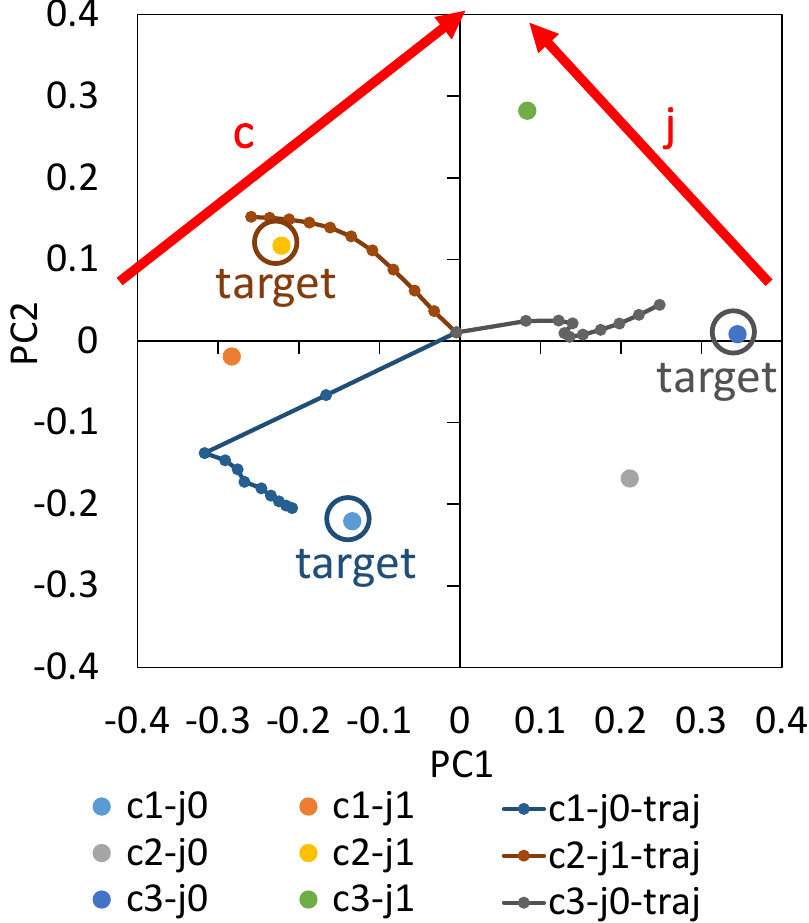}
  \caption{The trained parametric bias and its trajectory when conducting online update of parametric bias at c1-j0, c2-j1, and c3-j0.}
  \label{figure:parametric-bias}
\end{figure}

\begin{figure*}[t]
  \centering
  \includegraphics[width=1.9\columnwidth]{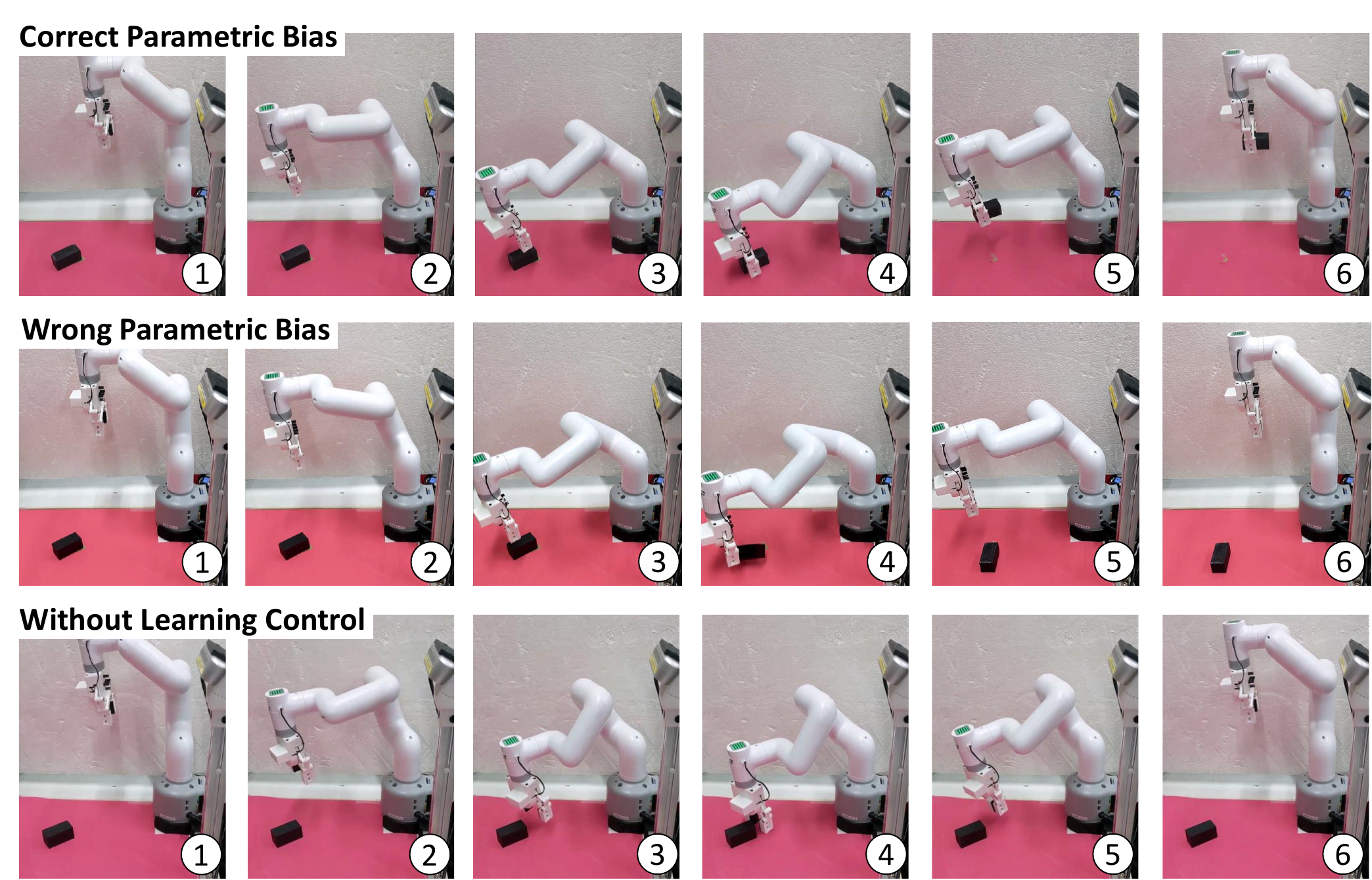}
  \caption{The experiment of visual servoing with correct or wrong parametric bias, or without any learning control.}
  \label{figure:vservoing-exp}
\end{figure*}

\begin{figure*}[t]
  \centering
  \includegraphics[width=1.99\columnwidth]{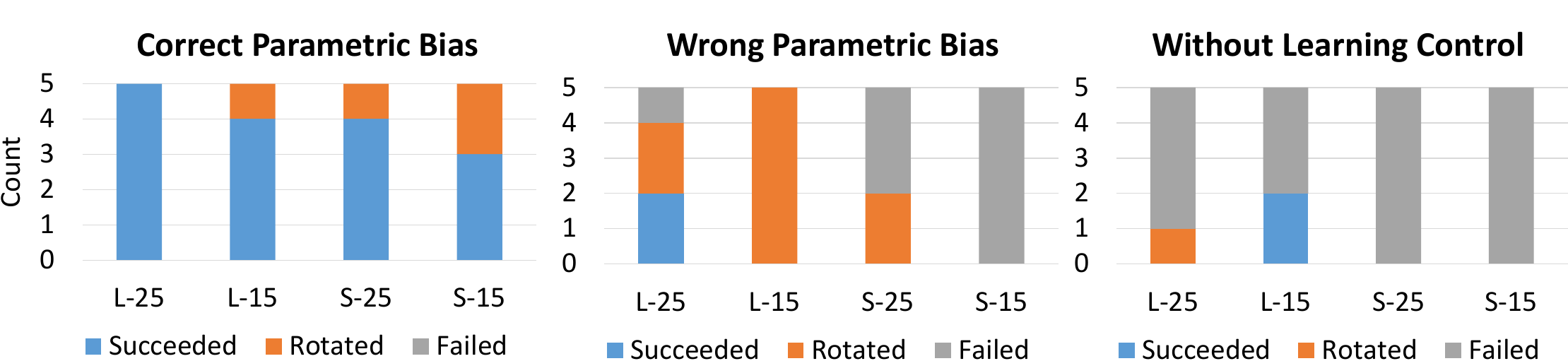}
  \caption{The ratio of Succeeded, Rotated, and Failed when conducting visual servoing with correct or wrong parametric bias, or without any learning control.}
  \label{figure:vservoing-eval}
\end{figure*}

\begin{figure}[t]
  \centering
  \includegraphics[width=0.7\columnwidth]{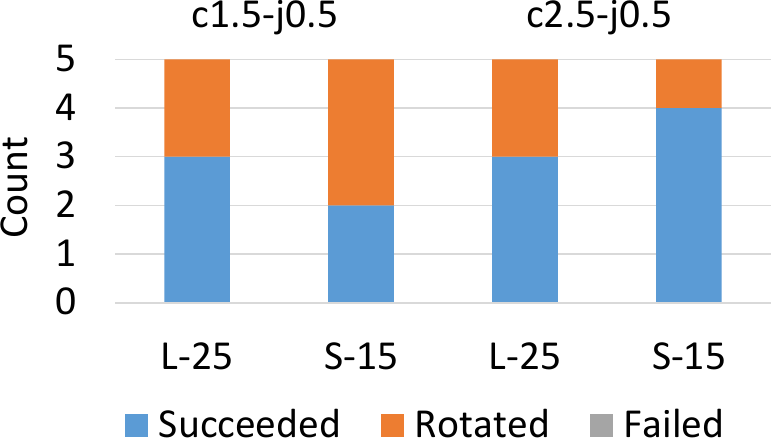}
  \caption{The ratio of Succeeded, Rotated, and Failed when conducting visual servoing with untrained body state and after online learning of PB.}
  \label{figure:vservoing-eval2}
\end{figure}

\subsection{Data Collection and Training Experiment}
\switchlanguage%
{%
  The data collection process is shown in \figref{figure:data-collection}.
  Note that (1)--(5) in \figref{figure:data-collection} denotes the numbers in \secref{subsec:data-collection}, and \ctext{1}--\ctext{14} denotes the image numbers.
  First, we set an initial posture and let the robot grasp an object in the desired way.
  The robot places the object at a random position within the range where inverse kinematics can be solved.
  In this case, it is necessary to reflect the desired motion during the actual grasping task to the object placement motion.
  In this study, we define the state of pregrasp at \ctext{4}, the robot moves quickly to this point, and then slowly lowers the hand to \ctext{5} to open the gripper.
  The robot then returns to the pregrasp state in the same manner at \ctext{6}, and from there returns to the initial posture.
  Next, the steps from \ctext{3} to \ctext{8} are performed in exactly the same manner except for the opening and closing of the gripper.
  The robot reaches its hand out to the exact same place as in \ctext{5}, makes the pregrasp posture at \ctext{10}, grasps the object at \ctext{11}, and returns to the initial posture.
  By repeating this procedure many times, it is possible for the robot to autonomously collect data.

  In this study, we train VSNPB by collecting data while changing the body state into six different states and performing 30 trials for each of the four types of objects (five trials for each body state), totaling 120 trials.
  Parametric bias $\bm{p}_{k}$ trained in this process is shown in \figref{figure:parametric-bias}.
  We can see that the parametric bias is regularly self-organized along the axes of \{c1, c2, c3\} and \{j0, j1\}.

  We also conducted an experiment in which the body state is set to \{c1-j0, c2-j1, c3-j0\} and the control in \secref{subsec:visual-servoing} is conducted while the online update in \secref{subsec:online-update} is performed simultaneously.
  The trajectory of the parametric bias $\bm{p}$ is shown as ``-traj'' in \figref{figure:parametric-bias}.
  It is found that the current parametric bias gradually approaches the parametric bias trained in the current body state, and that it is possible to accurately recognize the current body state from the prediction error of the sensors during the motion.
}%
{%
  データ収集の様子を\figref{figure:data-collection}に示す.
  なお, \figref{figure:data-collection}中の(1)--(5)は\secref{subsec:data-collection}における番号を表し, \ctext{1}--\ctext{14}は画像の番号を表す.
  まず初期姿勢を作り, 人間が把持して欲しい所望の形でロボットに物体を把持させる.
  逆運動学が解ける範囲でランダムな位置に物体を配置する.
  この際, 実際に把持する際の所望の動きを物体配置時にも反映させる必要がある.
  本研究では, \ctext{4}のPregraspの状態を定義し, ここまで素早く動き, そこから\ctext{5}にかけてゆっくりと体を下に下ろした後にグリッパを開く.
  そして, 同じように\ctext{6}でPregraspの状態まで戻り, そこから初期位置まで戻る.
  先の\ctext{3}から\ctext{8}までの動作をグリッパの開閉以外は全く同様に行う.
  \ctext{5}と全く同じ場所に手を伸ばしていき, \ctext{10}でPregrasp, \ctext{11}で把持を行い, 初期位置に戻る.
  この操作を何度も繰り返すことで, 自律的にロボットがデータを収集することが可能になる.

  本研究では, 身体状態を6種類に変化させながら, 4種類の物体それぞれについて30回ずつ(一つの身体状態につき5回ずつ), 計120回の試行を行いVSNPBを学習させた.
  この際に学習されたParametric Bias $\bm{p}_{k}$を\figref{figure:parametric-bias}に示す.
  Parametric Biasが\{c1, c2, c3\}と\{j0, j1\}の軸に沿って規則的に自己組織化していることがわかる.

  また, ロボット状態を\{c1-j0, c2-j1, c3-j0\}に設定して, \secref{subsec:visual-servoing}の制御と同時に\secref{subsec:online-update}を行う実験を行った.
  その際のParametric Bias $\bm{p}$の軌跡を\figref{figure:parametric-bias}の``-traj''として示す.
  現在のParametric Biasは, 現在のロボット状態において学習されたParametric Biasへと徐々に近づいていき, 動きの際の感覚の予測誤差から現在のロボット状態を正確に認識することが可能であることがわかった.
}%

\subsection{Visual Servoing Experiment}
\switchlanguage%
{%
  The experimental results of the actual visual servoing are shown in \figref{figure:vservoing-exp} and \figref{figure:vservoing-eval}.
  Four objects are placed at different positions and are grasped five times each by visual servoing.
  Here, "Succeeded" means that the robot succeeded in grasping the object, "Rotated" means that the robot succeeded in grasping the object but the gripper hit the edge and the object rotated, and "Failed" means that the robot failed to grasp the object.
  In all experiments, the body state is set to c2-j0, and we compare the case where the correct parametric bias trained for the same state is used (Correct) with the case where the wrong parametric bias trained for the state c3-j1 is used (Wrong).
  We also compare a general object grasping method that does not use VSNPB, such as detecting a tabletop object using plane detection and euclidean clustering, and solving inverse kinematics to it.
  Here, we use a rigid body approximation of the robot and the camera position on CAD (although it is possible to tune the robot and camera model over time, we use the geometric model as it is to show the control error of the low-rigidity robot).
  \figref{figure:vservoing-exp} shows examples of motions for the object S-25.
  The motion depends on the value of PB, and it can be seen that the robot correctly grasps the object when the value of PB is Correct, while it fails when the value of PB is Wrong.
  In the case without the learning control, the hand is greatly displaced and does not even touch the object.
  In \figref{figure:vservoing-eval}, we show the success rate for each object.
  In the case of Correct, all graspings Succeeded or Rotated, while in the case of Wrong, the probability of failure increases.
  This signifies that the recognition of the current body state is important.
  In both cases, the larger the object is, the higher the success rate of the grasping is.
  In the case of a small object, a small deviation often causes a failure of the grasp.
  When the learning control is not used, the grasping fails in most cases due to model errors.
  Some of the grasping is considered to be successful, but this is only the case when the robot can just barely grasp the edge of a long object.

  Finally, regarding the body states that were not used during the training, the results of the same experiment as above for L-25 and S-15 are shown in \figref{figure:vservoing-eval2} after the online update of PB.
  Here, the experiments are conducted for the cases where the body state is set as a combination of the following: c1.5 between c1 and c2 or c2.5 between c2 and c3 for the camera position, and j0.5 between j0 and j1 for the realization of joint angles, i.e. c1.5-j0.5 and c2.5-j0.5.
  The results show that the success rates for both cases are higher than those of Wrong cases in \figref{figure:vservoing-eval}, and not much different from those of Correct cases.
  We can see that even the body states that were not used in the training can be estimated by updating PB, and the visual servoing can be executed appropriately according to the body state.
}%
{%
  実際に視覚サーボを行った際の実験結果を\figref{figure:vservoing-exp}と\figref{figure:vservoing-eval}に示す.
  4つの物体をそれぞれ異なる位置に配置し, これを視覚サーボによりそれぞれ5回ずつ把持する実験を行う.
  ここで, Succeededは把持に成功した状態, Rotatedは把持には成功しているがグリッパが物体の端に当たり物体が回転してしまった状態, Failedは物体把持に失敗した状態を表す.
  全ての実験でロボット状態はc2-j0とし, 同様の状態について学習されたParametric Biasを使った場合(Correct)と, c3-j1の状態について学習された間違ったParametric Biasを使った場合(Wrong)を比較している.
  また, 平面検出から卓上の物体をユークリッドクラスタリングにより検知し, これに対してグリッパをアプローチするようなVSNPBを用いない一般的な物体把持手法についても比較している.
  この際は剛体近似されたロボットとCAD上のカメラ位置等を用いている(時間をかけてチューニングをすることも可能あるが, 低剛性ロボットの誤差を示すために幾何モデルをそのまま使用している).
  \figref{figure:vservoing-exp}は物体S-25における動作例であるが, PBの値によって動作が異なり, Correctの場合は正しく把持できているのに対して, Wrongの場合は失敗してしまっていることがわかる.
  また, 学習制御を用いない場合は大きくハンドの位置がズレて, 物体に触れることすら出来ていない.
  \figref{figure:vservoing-eval}には, それぞれの物体に対する成功率を表示している.
  Correctの場合はSucceededまたはRotatedで全て把持に成功しているのに対して, Wrongの場合は, 把持に失敗する確率が高まっている.
  つまり, 現在のロボット状態の認識が重要であることがわかる.
  またどちらの場合も, 物体が大きい方がより把持の成功率が高いことがわかる.
  小さな物体は多少のズレで把持が失敗してしまうことが多い.
  一方, 学習制御を用いない場合はほとんどの場合でモデル誤差により物体把持に失敗している.
  一部把持に成功していることもあるが, 長い物体の端をギリギリ掴めた程度である.

  最後に, 学習時にデータとして用いなかったロボット状態において, PBのオンライン学習実行後に先と同様の実験をL-25とS-15について行った際の結果を\figref{figure:vservoing-eval2}に示す.
  ここでは, ロボット状態を関節角度の実現度をj0とj1の間のj0.5, カメラの位置をc1とc2の間のc1.5またはc2とc3の間のc2.5とした場合について実験を行っている.
  結果は, 両者とも\figref{figure:vservoing-eval}のWrongの場合よりも成功率が高く, Correctの場合と大差ない結果となった.
  つまり, 訓練時に用いなかったロボット状態でも, PBの更新によりこれを推定し, 適切に視覚サーボを実行することができることがわかった.
}%

\section{Discussion} \label{sec:discussion}
\switchlanguage%
{%
  We discuss the results of our experiments.
  First, regarding data collection, we found that even a low-rigidity robot can autonomously collect data by using the reproducibility of its own body movements.
  On the other hand, there are a few parts that are not reproducible.
  If we want to perform 10 trials, there is no problem, but in order to perform 100 trials, a manual adjustment of the grasping position of the object is necessary halfway through the trials.
  Next, in the training of VSNPB, the parametric bias self-organizes regularly along the body state change.
  In addition, online learning enables us to recognize the current body state accurately.
  Finally, for the actual visual servoing, though the success rate varies depending on the size of the object, the grasping is successful with a high probability if the body state is correctly recognized.
  Even if the body state is not given as data for training, it is possible to perform the grasping accurately by combining with the online update of PB.
  On the other hand, the failure rate of grasping increases when the body state is not recognized well, and there are many cases where the robot cannot even touch the object without the learning control.

  A major advantage of this study is the autonomy of the robot to collect data by itself.
  In addition, for example, if data is collected for each state in a robot whose screws are gradually loosening, it will be possible to accurately handle the body state of the robot both when the screws are tightened and when they are loosened.
  It is also important to note that we do not need to reproduce a low-rigidity body in a simulation, nor do humans need to teach the robot how to move.
  Also, as this study does not assume a specific body structure, it can be applied to a variety of robots such as musculoskeletal humanoids \cite{kawaharazuka2019musashi} or flexible octopus-like robots \cite{laschi2012octopus} in the future.
  In this study, we chose simple rectangular objects as objects to be grasped to emphasize the concept of the method, but we believe that we can show the advantages of this learning control further by applying it to more complex objects.

  Finally, we discuss the limitations of this study.
  First, our system assumes that the object is placed on a flat surface when it is randomly placed, and that the robot can grasp the object if it behaves exactly as it did when placing the object.
  Therefore, if the environment in which the objects are placed is complex and the position and angle of the object changes at the moment of placing the object, the robot cannot grasp the object even if it reaches to the same place.
  Secondly, this study does not deal with multiple objects, and this problem also needs to be solved.
  For this, we need to label the objects to be grasped in the network input, or use implicit expressions as in Implicit Behavior Cloning \cite{florence2021implicit}.
}%
{%
  これまでの実験から考察を述べる.
  まず, データ収集についてであるが, 低剛性なロボットでも自身の身体動作の再現性を用いることで, 自律的にデータを収集することが可能であるとわかった.
  一方で, 微小ではあるが再現性の取れない部分もあり, 10回程度であれば問題はなかったが, 100回程度の試行を行いたい場合は現状途中で人間が物体の把持位置を調整してあげる必要がある.
  次にVSNPBの学習であるが, Parametric Biasはその身体状態変化に沿って規則的に自己組織化した.
  また, オンライン学習により現在の身体状態についても正確に認識することができる.
  最後に実際の視覚サーボについてであるが, 物体の大きさによって成功率にばらつきはあるものの, 身体状態が正しく認識されていれば, 高い確率で把持に成功することがわかった.
  訓練時にデータとして与えていない身体状態でも, PBのオンライン学習と合わせることで正確に把持を実行することができる.
  一方で, 身体状態の認識が上手く行っていない場合は把持の失敗率が高まり, 本学習制御を用いない場合は物体に触れることすらできない場合が多かった.

  本研究の大きな利点はデータを勝手に収集してくれる自律性にある.
  また, 例えば徐々にネジが緩むようなロボットにおいて, それぞれの状態についてデータをとっておけば, ネジを締め直したばかりの状態と, その後緩んでいった状態の両方の身体状態を正確に扱うことができるようになる.
  シミュレーション上に低剛性な身体を再現したり, 人間が動きを教えたりする必要がなく, ロボットとカメラさえあれば実行できる手軽さも重要である.
  また, 身体構造を仮定しないため, 筋骨格ヒューマノイド\cite{kawaharazuka2019musashi}やタコのような柔軟ロボット\cite{laschi2012octopus}等への応用も今後進むと考えられる.
  本研究では把持物体として単純な直方体を選び手法のコンセプトを強調したが, より複雑な把持物体へと応用していくことで, より学習型制御の良さを表現できると考える.

  最後に, 本研究の限界について述べる.
  まず, 本研究はランダムに物体を置く際に平面に置くことを仮定しており, 物体配置時と全く同じように動作すれば物体を把持することができるという強い仮定が入っている.
  そのため, もし物体を配置する環境がデコボコであり, 物体配置した瞬間に物体の傾きや角度が変化してしまうような場合, 同じ場所に手を伸ばしても把持できないため上手く動作しないと考えられる.
  次に, 本研究は複数物体を扱っておらず, この問題も解決する必要がある.
  これはネットワークの入力に把持する物体のラベルを入れる, またはImplicit Behavior Cloning \cite{florence2021implicit}のような陰的な表現を用いる必要があると考える.
}%

\section{CONCLUSION} \label{sec:conclusion}
\switchlanguage%
{%
  In this study, we proposed a deep learning model to handle automatic data collection and temporal body changes, which are problematic when a low-rigidity robot performs visual servoing.
  For data collection, we proposed a method where the robot autonomously collects data by repeatedly placing an object and reaching out to the same place by itself, taking advantage of the fact that the low-rigidity robot has difficulty performing the intended movement but has reproducibility in motion.
  For the temporal body changes, we introduced parametric bias, which can embed the implicit changes in dynamics into the model, and proposed a method for the robot to adapt to its current body state by updating the parametric bias to match the current prediction and the actual measurement.
  We applied this method to an actual robot with low-rigidity, and confirmed its effectiveness for a grasping task for several objects with various positions and angles.
  In the future, we would like to construct a system that can autonomously learn the relationship between body sensors and motions even for low-rigidity, flexible, and complex robots, taking into account multiple target objects and various ways of grasping.
}%
{%
  本研究では, 低剛性な実機ロボットが視覚サーボを行う際に問題となるデータ収集と身体状態変化に対応する深層学習モデルの提案を行った.
  データ収集については, 低剛性のため意図した動きは難しい一方でその動きに再現性があることを利用し, 自身で物体を置き同じところに手を伸ばすことを繰り返すことで自律的にデータ収集を行うことを提案した.
  身体状態変化については, 暗黙的なダイナミクスの変化を埋め込むことが可能なParametric Biasを導入し, 現在の予測と実測を合致させるようにこれを更新することで現在の身体状態に適応する方法を提案した.
  これを低剛性な実機ロボットに実際に適用し, 数種類の物体の位置と角度に対してサーボを行うような系について, その有効性を確認した.
  今後は, 複数物体や物体を置く角度等を考慮し, 低剛性で柔軟かつ複雑なロボットでも自律的に身体の感覚と運動の関係を学習するシステムの構築を行いたい.
}%

{
  \bibliographystyle{IEEEtran}
  \bibliography{main}
}

\end{document}